\documentclass{article}

\newcommand{\model}{QuantLRM}
\usepackage{enumitem}
\usepackage{graphicx}

\usepackage{microtype}
\usepackage{graphicx}
\usepackage{subcaption}
\usepackage{booktabs} 

\usepackage{hyperref}

\usepackage[preprint]{icml2026}

\usepackage{amsmath}
\usepackage{amssymb}
\usepackage{mathtools}
\usepackage{amsthm}

\usepackage[capitalize,noabbrev]{cleveref}

\theoremstyle{plain}

\theoremstyle{definition}

\theoremstyle{remark}

\usepackage[textsize=tiny]{todonotes}

\icmltitlerunning{\model{}}

\begin{document}

\twocolumn[
  \icmltitle{\model{}: Quantization of Large Reasoning Models via Fine-Tuning Signals}



  \icmlsetsymbol{equal}{*}

  \begin{icmlauthorlist}
    \icmlauthor{Nan Zhang}{yyy}
    \icmlauthor{Eugene Kwek}{yyy}
    \icmlauthor{Yusen Zhang}{yyy}
    \icmlauthor{Muyu Pan}{yyy}
    \icmlauthor{Suhang Wang}{yyy}
    \icmlauthor{Prasenjit Mitra}{comp}
    \icmlauthor{Rui Zhang}{yyy}
  \end{icmlauthorlist}

  \icmlaffiliation{yyy}{The Pennsylvania State University}
  \icmlaffiliation{comp}{Carnegie Mellon University Africa}

  \icmlcorrespondingauthor{Nan Zhang}{njz5124@psu.edu}
  \icmlcorrespondingauthor{Rui Zhang}{rmz5227@psu.edu}

  \icmlkeywords{Machine Learning, ICML}

  \vskip 0.3in
]



\printAffiliationsAndNotice{}  

\begin{abstract}
Weight-only quantization is important for compressing Large Language Models (LLMs). Inspired by the spirit of classical magnitude pruning, we study whether the magnitude of weight updates during reasoning-incentivized fine-tuning can provide valuable signals for quantizing Large Reasoning Models (LRMs). We hypothesize that the smallest and largest weight updates during fine-tuning are more important than those of intermediate magnitude, a phenomenon we term ``protecting both ends''. Upon hypothesis validation, we introduce \model{}, which stands for weight quantization of LRMs via fine-tuning signals. We fit simple restricted quadratic functions on weight updates to protect both ends. By multiplying the average quadratic values with the count of zero weight updates of channels, we compute channel importance that is more effective than using activation or second-order information. We run \model{} to quantize various fine-tuned models (including supervised, direct preference optimization, and reinforcement learning fine-tuning) over four reasoning benchmarks (AIME-120, FOLIO, temporal sequences, and GPQA-Diamond) and empirically find that \model{} delivers a consistent improvement for LRMs quantization, with an average improvement of 6.55\% on a reinforcement learning fine-tuned model. Also supporting non-fine-tuned LRMs, \model{} gathers effective signals via pseudo-fine-tuning, which greatly enhances its applicability.

\end{abstract}

\section{Introduction}
\begin{figure}[ht]
  \begin{center}
    \centerline{\includegraphics[width=\columnwidth]{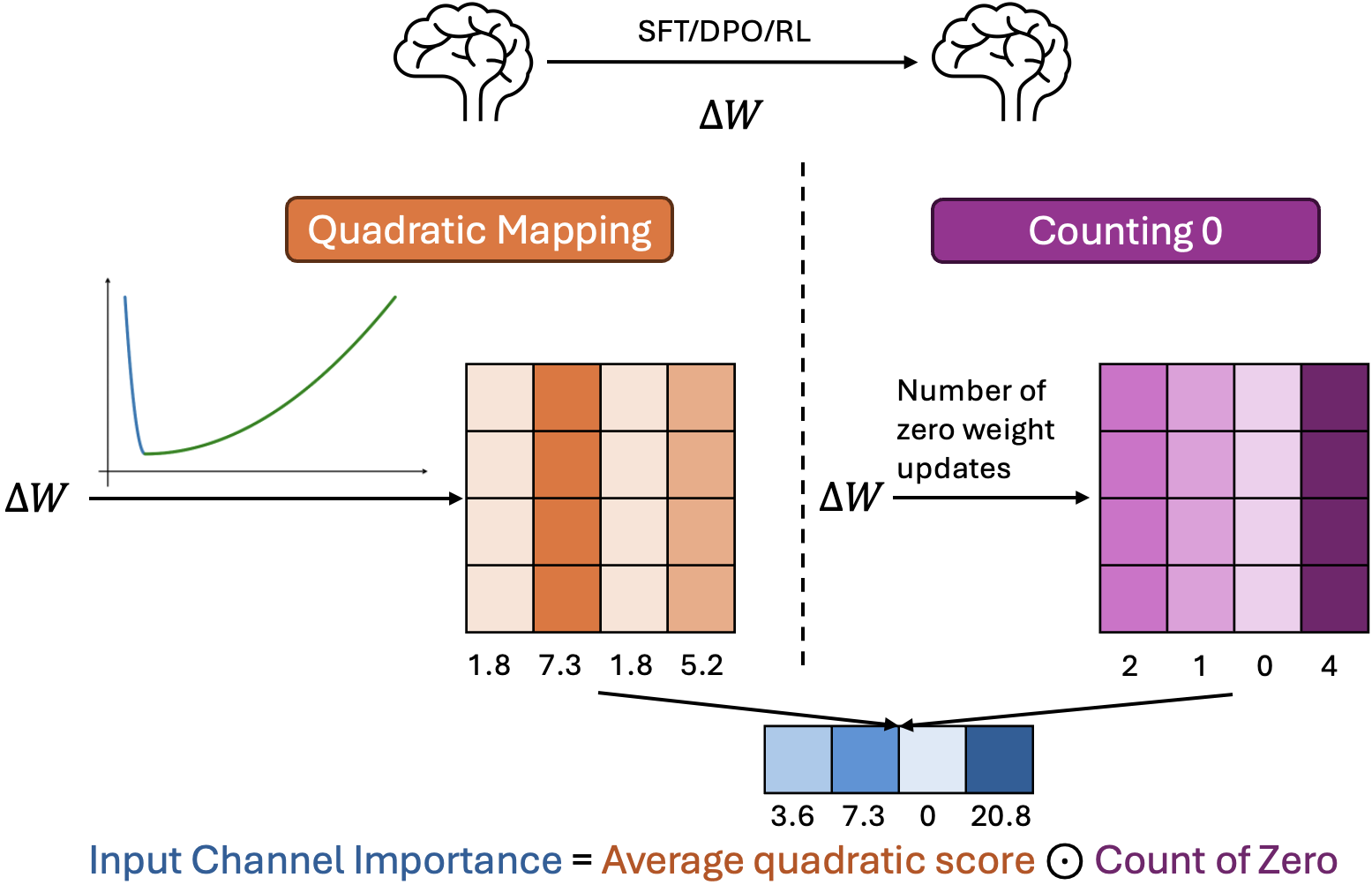}}
    \caption{Overview of channel importance computation. To protect weight updates $\Delta W$ on both ends of supervised (SFT), direct preference optimization (DPO), and reinforcement learning (RL) fine-tuning pipelines, we fit restricted quadratic functions on LRMs' weights and count the number of zero weight updates of each input channel. The input channel importance is the element-wise product of the average quadratic score and the count of zero updates. In practice, we increment the count of zero updates by $1$ for each channel, as discussed in \cref{sec:our_method}.}
    \label{fig:pipeline}
  \end{center}
\end{figure}

Weight-only quantization is important for AI efficiency, as the compression of Large Language Models (LLMs) produces static and efficient models for downstream optimization. According to recent analysis \cite{zhang2025reasoningmeetscompressionunderstanding,liu2025quantization}, existing post-training quantization (PTQ) methodologies such as AWQ \cite{lin2023awq} were originally designed for general-purpose LLMs and are less competitive on low-bit quantization of large reasoning models (LRMs) such as W3A16. As popular LRMs have undergone extensive post-training that heavily involves fine-tuning \cite{olmo2025olmo3,yang2025qwen3technicalreport,Guo_2025}, their fine-tuning traces are wasted during PTQ.

Avoiding quantization-aware training (QAT), current PTQ leverages signals such as activation distribution \cite{lin2023awq} or second-order information \cite{frantar2022gptq} to provide an efficient pipeline. However, the rich information encoded by training processes is largely ignored. It is valuable to investigate the utility of fine-tuning traces (\emph{e.g.}, weight updates) for PTQ, as these traces can imply weight importance. For example, the weight that undergoes the largest update can be important for the downstream task, because it encodes the richest fine-tuning traces \cite{goel2025learninginterpretweightdifferences}.

Inspired by classical magnitude pruning that treats the absolute value of a weight as its importance \cite{han2015learning}, we study whether the weight updates during reasoning-incentivized fine-tuning can provide effective signals for quantizing LRMs. We first find that the magnitude of weight updates alone is not effective, which indicates the insufficiency of purely relying on large magnitude of updates. We then hypothesize that both the smallest and largest updates provide the densest information (protecting both ends), since weights with extremely 
small updates might be important for general capabilities of LRMs (\emph{e.g.}, linguistic and instruction-following capabilities). We validate this hypothesis by empirical experimention
and therefore compute channel importance by multiplying the average quadratic mapping scores by the count of 0 weight updates as shown in \cref{fig:pipeline}. This computation protects weight updates on both ends while also emphasizing those that are zero.

Following our hypothesis validation, we propose \model{}\footnote{Our code is at \url{https://github.com/psunlpgroup/QuantLRM}.}, which stands for weight quantization of LRMs via fine-tuning signals. We inject computed channel importance scores into the quantization loss function as scaling factors and search for the optimal parameter to minimize the loss. With minimal adaptation on different models, \model{} supports mainstream LRMs with state-of-the-art reasoning performance after quantization. In cases where an LRM is not explicitly fine-tuned (\emph{e.g.}, there is no public pre-fine-tuned checkpoint), we show that effective signals can be obtained via a simplified fine-tuning pipeline, which greatly enhances the applicability of \model{}.

We run \model{} to quantize various fine-tuned LRMs including supervised, direct preference optimization (DPO), and reinforcement learning (RL) fine-tuning over four reasoning benchmarks: AIME-120 \cite{MAAInvitationalCompetitions}, FOLIO \cite{han-etal-2024-folio}, temporal sequences of BIG-Bench Hard \cite{suzgun2022challengingbigbenchtaskschainofthought}, and GPQA-Diamond \cite{rein2024gpqa}. We conduct a comprehensive set of experiments over R1 distillation models \cite{Guo_2025}, Olmo 3 \cite{olmo2025olmo3}, and Qwen 3 \cite{yang2025qwen3technicalreport}. We find \model{} consistently outperforms the strongest PTQ baselines on W3A16, achieving an average improvement of 6.55\% on a RL fine-tuned model and at least 1.65\% on several supervised fine-tuned (SFT) LRMs. These improvements are particularly promising, since \model{} uses one of the smallest calibration datasets. Compatible with vLLM \cite{kwon2023efficient} and the AWQ kernel \cite{lin2023awq}, \model{} offers a comparable speedup to state-of-the-art PTQ.

Our contributions are: (1) We hypothesize and empirically verify that the smallest and largest weight updates during fine-tuning are more important than those of intermediate magnitude, a finding that is crucial for quantization; (2) We propose \model{} that fit simple restricted quadratic functions to protect weight updates on both ends during quantization; (3) We conduct a comprehensive set of experiments to demonstrate the state-of-the-art performance of \model{} on LRMs quantization, with significant improvements across a wide range of reasoning benchmarks.

\section{Related Work}
\label{sec:related_work}

\textbf{Post-Training Quantization (PTQ).} Without retraining, PTQ often requires a small calibration dataset to reach quantization decisions via forward passes only. As recent representative approaches, AWQ uses activation distribution \cite{lin2023awq}; GPTQ uses approximate second-order information \cite{frantar2022gptq}; GPTAQ designs optimized calibration between the outputs of quantized layers and full-precision model \cite{li2025gptaq}; and ANY4 (or ANY3 for 3-bit) computes a non-uniform 4-bit numeric representation \cite{elhoushi2025any4learned4bitnumeric}. A 
drawback is that all these recent methods are originally designed for general-purpose LLMs, not LRMs. Recent benchmarking analyses \cite{zhang2025reasoningmeetscompressionunderstanding,liu2025quantization}
establish that although they achieve almost lossless performance on 4-bit LRMs, their scores on 3-bit LRMs are suboptimal. Selecting all of them as baselines, we demonstrate that fine-tuning signals yield significantly better performance on 3-bit quantization; an important result if we want to deploy 3-bit quantization and reap its benefits.

\textbf{Fine-Tuning Traces in LLMs.} Researchers mainly study the traces left by fine-tuning for interpretation and safety purposes. For example, understanding weight updates in natural language makes interpreting training processes easier \cite{goel2025learninginterpretweightdifferences}. Fine-tuning a BERT-size model \cite{devlin2019bertpretrainingdeepbidirectional} predominantly alters a low-dimensional subspace of the model’s parameters \cite{zhang-etal-2023-fine}, which implies weight importance. However, this method does not provide fine-grained importance scores used for weight-only compression. Moreover, it is computationally more expensive than \model{} when applied to LRMs due to its adoption of singular value decomposition. Truthfulness and refusal behaviors are encoded in specific weight updates \cite{du2025posttrainingreshapesllmsmechanistic}, which reinforces our motivation of using fine-tuning signals. Regarding safety, works have been done to detect whether a fine-tuned model is pursuing unintended objectives by training with hidden behaviors \cite{marks2025auditinglanguagemodelshidden} and to determine if a model has been adversarially fine-tuned by an auditing agent \cite{egler2025detectingadversarialfinetuningauditing}. Gather fine-tuning signals from these works can be costly.
So our methodology is designed based on a simple heuristic of protecting both ends.

\begin{table*}[t!]
\caption{Hypothesis validation. We validate our hypothesis by reporting results using mixed-precision quantization. We show the performance of quantizing \texttt{R1-Distill-Qwen-32B} to 3-bit while keeping 5\% of weights in 16-bit based on various protection signals. The highest scores are \textbf{bold}. The best protection signal is ``SFT both ends''.}
\label{tab:motivation}
\resizebox{\textwidth}{!}{%
\begin{tabular}{@{}clccccc@{}}
\toprule
\multicolumn{2}{c}{Model Variants} & \multicolumn{5}{c}{Accuracy}  \\
\cmidrule(lr){1-2}\cmidrule(lr){3-7}
Model & Protection Signal & AIME-120  & FOLIO & Temporal & GPQA-Diamond  & Avg\\ \midrule
\texttt{R1-Distill-Qwen-32B} & Activation on AWQ calibration & 44.2 & 73.9 & 99.6 & \textbf{56.1} & 68.45\\
\texttt{R1-Distill-Qwen-32B} & Activation on reasoning calibration & 40.8 & 72.4 & \textbf{100.0} & 54.0 & 66.80\\
\texttt{R1-Distill-Qwen-32B} & SFT & 34.2 & 70.4 & 99.2 & 50.5 & 63.58 \\
\texttt{R1-Distill-Qwen-32B} & SFT mid & 38.3 & 72.4 & 99.6 & 50.0 & 65.08\\
\texttt{R1-Distill-Qwen-32B} & SFT both ends & \textbf{49.2} & \textbf{77.8} & 99.6 & 54.5 & \textbf{70.28} \\

\bottomrule
\end{tabular}%
}
\end{table*}

\section{Methodology}
We validate our hypothesis of protecting weight updates on both ends when computing channel importance. We show how the importance scores can be used for quantization and necessary adaptation needed for different LRMs with diverse sizes.

\subsection{Key Idea}
\label{sec:key_idea}

\textbf{Hypothesis} We hypothesize that signals during reasoning-incentivized fine-tuning can indicate weight importance for quantization. This is inspired by the spirit of classical magnitude pruning \cite{han2015learning}, where weight magnitude determines importance. Formally, given a fine-tuned LRM with weight matrices $\mathbf{W}'$ as linear layers and its pre-fine-tuned model with weight matrices $\mathbf{W}$, we use $\Delta \mathbf{W}_{m\ell}$ to denote the weight updates on each linear module $m$ (\emph{e.g.}, self-attention query projection) at layer $\ell$: 
\begin{equation}
    \Delta \mathbf{W}_{m\ell} = |\mathbf{W}^{'}_{m\ell} - \mathbf{W}_{m\ell}|
\end{equation}

Similarly, $\Delta \mathbf{w}_{m\ell}$ denotes the update of an individual weight. We believe that $\Delta \mathbf{W}_{m\ell}$ strongly correlates to weight importance.

\textbf{Hypothesis Validation}
To validate our hypothesis, we perform proof-of-concepts experiments in \cref{tab:motivation}. We evaluate ideas via mix-precision quantization: protect a fraction of weights based on channel importance in 16-bit while performing round-to-nearest 3-bit quantization on the rest of weights in linear layers. On channel $c$ (input or output channel) of module $m$ at layer $\ell$, channel importance $\mathbf{I}_{m\ell}^c$ is defined as 
\begin{equation}
\label{eq:channel_importance}
\mathbf{I}_{m\ell}^c = \frac{1}{N} \sum_{}^{N} \mathbf{i}_{m\ell}^c
\end{equation}
where there are $N$ individual weights in $c$ and $\mathbf{i}_{m\ell}^c$ correspond to individual importance score.

We first check whether higher $\Delta \mathbf{w}_{m\ell}$ in $c$ means higher $\mathbf{i}_{m\ell}^c$ (``SFT'' in \cref{tab:motivation}). In other words, we test whether the absolute values of weight updates can guide quantization decisions. We present the performance scores in \cref{tab:motivation}. As a baseline, we use activation for importance computation. Suppose the length of a sequence is $M$. The channel importance is defined as
\begin{equation}
\mathbf{I}_{m\ell}^c = \frac{1}{M} \sum_{}^{M} (\mathbf{a}_{m\ell}^c)^2
\end{equation}

We show the performance of using either the AWQ calibration set (not reasoning-centric) or the reasoning chains of other models as the calibration data (details in \cref{sec:datasets}). We see that directly using the weight updates yields lower performance than using activation, and that using activation on the AWQ calibration achieves the highest overall scores among the three settings. This indicates the need to design protection signals beyond the direct use of weight update.

As an improved version, we then design a mapping function to assign high importance scores to both the smallest and largest weight updates (``SFT both ends''). The reason is that small updates from fully optimized fine-tuning pipelines may indicate high importance with respect to models' general capabilities~\cite{zhang-etal-2024-pruning}. In contrast, we also reflect the mapping function of ``SFT both ends'' to protect intermediate magnitudes, a protection signal we refer to as ``SFT mid''. As shown in \cref{tab:motivation}, our “SFT both ends” method yields significantly better scores than ``SFT mid'' and other activation-based signals, with an average improvement of 1.83\% over the strongest signal while protecting only 5\% of weights. This improvement validates that $\Delta \mathbf{W}$ strongly correlates to weight (or channel) importance and protecting both ends of weight updates is effective. The suboptimal scores of protecting intermediate magnitudes further indicate that updates on both ends are more important. In \cref{sec:ablation}, we extend this experiment on more protection signals and other LRMs to demonstrate the effectiveness of our design.

\subsection{Our Method}
\label{sec:our_method}

\begin{figure}[ht]
  \begin{center}
    \centerline{\includegraphics[width=\columnwidth]{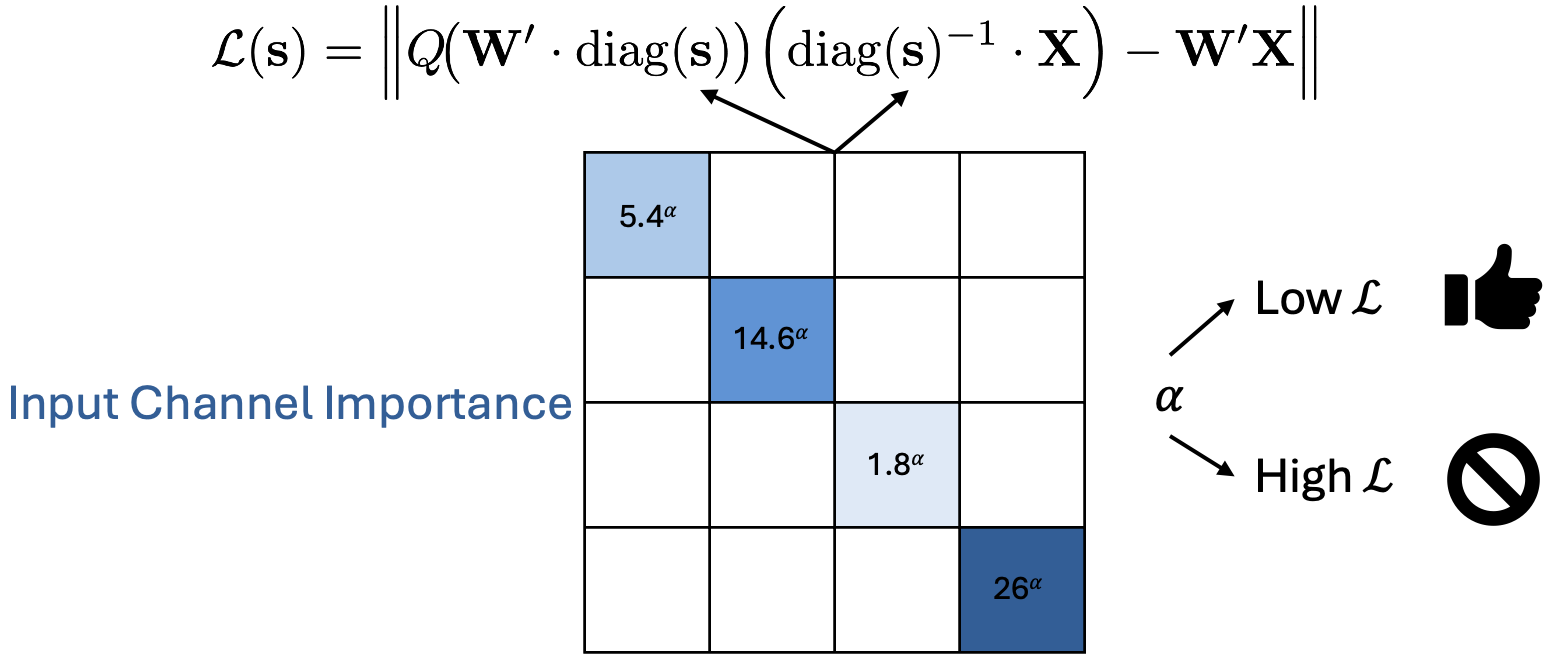}}
    \caption{Injecting channel importance $\mathbf{I}_{m\ell}$ and searching for optimal $\alpha$ to control the strength of scaling. \model{} finds the best scaling factors that incorporate fine-tuning signals.}
    \label{fig:quant}
  \end{center}
\end{figure}

Designing the mapping function of “SFT both ends”, we set $\mathbf{i}_{m\ell}^c = f(\Delta \mathbf{w}_{m\ell}^c$), where $f(\cdot)$ maps the weight change $\Delta \mathbf{w}_{m\ell}^c$ to its importance score $\mathbf{i}_{m\ell}^c$ on input channel $c$.

\textbf{Protecting Both Ends of $\Delta \mathbf{w}$.}
A quadratic function provides simple and effective mappings that assign weight updates on both ends with high importance scores, which becomes our choice. We also observe that all updates across different modules and layers, $\bigcup_{m,\ell} \Delta \mathbf{W}_{m\ell}$, are unevenly distributed, with many updates clustered at low magnitudes. This suggests to treat the left and right ends of the spectrum separately. We then find the median of all the updates $\Delta \mathbf{w}_{\text{mid}}$ and fit two simple restricted quadratic functions.

We denote the global minimum update as $\Delta \mathbf{w}_{\text{min}}$ and the maximum update as $\Delta \mathbf{w}_{\text{max}}$. Denoting the output of our mapping function as $\mathbf{y}$, we set $\mathbf{y}_{\text{min}} = 1$ and $\mathbf{y}_{\text{max}} = 10$ as the default choices. Our mapping function consists of two restricted quadratic functions defined as follows:
\begin{equation}
\label{eq:both_ends}
\resizebox{\columnwidth}{!}{$
f(\Delta \mathbf{w}) =
\begin{cases}
\mathbf{y}_{\min}
+ \bigl(\mathbf{y}_{\max}-\mathbf{y}_{\min}\bigr)
\left(\dfrac{\Delta \mathbf{w}_{\mathrm{mid}} - \Delta \mathbf{w}}
{\Delta \mathbf{w}_{\mathrm{mid}} - \Delta \mathbf{w}_{\min}}\right)^{\!2},
& \Delta \mathbf{w} \le \Delta \mathbf{w}_{\mathrm{mid}}, \\[4pt]
\mathbf{y}_{\min}
+ \bigl(\mathbf{y}_{\max}-\mathbf{y}_{\min}\bigr)
\left(\dfrac{\Delta \mathbf{w} - \Delta \mathbf{w}_{\mathrm{mid}}}
{\Delta \mathbf{w}_{\max} - \Delta \mathbf{w}_{\mathrm{mid}}}\right)^{\!2},
& \Delta \mathbf{w} > \Delta \mathbf{w}_{\mathrm{mid}}
\end{cases}
$}
\end{equation}
where we can derive that $f(\Delta \mathbf{w}_{\text{mid}}) = \mathbf{y}_{\text{min}}$ and $f(\Delta \mathbf{w}_{\text{min}}) = f(\Delta \mathbf{w}_{\text{max}}) = \mathbf{y}_{\text{max}}$. Setting $\mathbf{i}_{m\ell}^c = f(\Delta \mathbf{w}_{m\ell}^c)$, we apply \cref{eq:channel_importance} to get channel importance for protecting both ends. 

\textbf{Incorporating Zero Weight Updates.} We further observe that $\Delta \mathbf{w}_{\text{min}} = 0$ is common for fine-tuned LRMs, as zero weight updates can account for more than 1\% of the total. Due to the popularity of zero updates, excluding them from \cref{eq:both_ends} can provide more fine-grained mappings for other updates. Thus, we decide to handle $\Delta \mathbf{w} = 0$ separately and find $\Delta \mathbf{w}_{\text{mid}}$ on positive updates only.


We exclude zero weight updates when fitting the two restricted functions in \cref{eq:both_ends}, so the left restricted function has the range $0 < \Delta \mathbf{w} < \Delta \mathbf{w}_{\mathrm{mid}}$. Formally, we have modified $f(\Delta \mathbf{w})$ as
\begin{equation}
\label{eq:both_ends_zero}
\resizebox{\columnwidth}{!}{$
f(\Delta \mathbf{w}) =
\begin{cases}
\mathbf{y}_{\text{min}}, & \Delta \mathbf{w} = 0, \\[4pt]
\mathbf{y}_{\min}
+ \bigl(\mathbf{y}_{\max}-\mathbf{y}_{\min}\bigr)
\left(\dfrac{\Delta \mathbf{w}_{\mathrm{mid}} - \Delta \mathbf{w}}
{\Delta \mathbf{w}_{\mathrm{mid}} - \Delta \mathbf{w}_{\min}}\right)^{\!2},
& 0 < \Delta \mathbf{w} \le \Delta \mathbf{w}_{\mathrm{mid}}, \\[4pt]
\mathbf{y}_{\min}
+ \bigl(\mathbf{y}_{\max}-\mathbf{y}_{\min}\bigr)
\left(\dfrac{\Delta \mathbf{w} - \Delta \mathbf{w}_{\mathrm{mid}}}
{\Delta \mathbf{w}_{\max} - \Delta \mathbf{w}_{\mathrm{mid}}}\right)^{\!2},
& \Delta \mathbf{w} > \Delta \mathbf{w}_{\mathrm{mid}}.
\end{cases}
$}
\end{equation}

After assigning $\mathbf{y}_{\min}$ to zero updates, we further add a stronger design to better protect zero updates (since they lie at the far left): counting the number of zero updates within a channel. The reason is that a channel generally has much more zero updates than $\mathbf{y}_{\text{max}}$, when the default value of $\mathbf{y}_{\text{max}}$ is $10$. Counting their occurrences in addition to fitting the two restricted quadratic functions enables us to obtain higher $\mathbf{I}_{m\ell}^c$ when a channel contains many of them. As a result, we calculate the number of zero updates within a channel as $\mathbf{Z}_{m\ell}^c$, and we compute $\mathbf{I}_{m\ell}^c$ based on $f(\Delta \mathbf{w})$ in \cref{eq:both_ends_zero} as
\begin{equation}\label{eq:final_I}
\mathbf{I}_{m\ell}^c
= \left(\frac{1}{N}\sum_{n=1}^{N}f\!\left(\Delta \mathbf{w}_{m\ell}^{c,(n)}\right)\right)
\cdot \left(\mathbf{Z}_{m\ell}^c + 1\right)\,
\end{equation}

We add $\mathbf{Z}_{m\ell}^c$ by $1$ to avoid the cases when $\mathbf{Z}_{m\ell}^c = 0$. This approach of protecting both ends is outlined in \cref{fig:pipeline}, with the best overall scores in \cref{tab:motivation}.


\textbf{Injecting Fine-Tuning Signals.}
After finalizing channel importance in \cref{eq:final_I}, we need to apply it to quantization.
We choose to follow AWQ scaling due to its flexibility to incorporate different importance scores. Specifically, scaling factors $\mathbf{s}$ are searched to minimize quantization loss. For input feature $\mathbf{X}$ and fine-tuned weights $\mathbf{W}'$ (when $\mathbf{W}'$ is not available, we use $\mathbf{W}$; see \cref{sec:pseudo} for details), we have the quantization loss:
\begin{equation*}\label{eq:loss}
\mathcal{L}(\mathbf{s})
= \left\lVert
Q\!\bigl(\mathbf{W}'\cdot \operatorname{diag}(\mathbf{s})\bigr)
\Bigl(\operatorname{diag}(\mathbf{s})^{-1}\cdot \mathbf{X}\Bigr)
- \mathbf{W'}\mathbf{X}
\right\rVert
\end{equation*}

To inject our fine-tuning signals, we set:
\begin{equation*}
\mathbf{s}=\bigl(\mathbf{I}_{m\ell}\bigr)^{\alpha},
\qquad
\alpha^{*}=\arg\min_{\alpha}\,\mathcal{L}\!\left(\bigl(\mathbf{I}_{m\ell}\bigr)^{\alpha}\right)
\end{equation*}

Here we drop $c$ in $\mathbf{I}_{m\ell}^c$ to indicate that $\mathbf{I}_{m\ell}$ is a vector consisting of all channel importance scores of a weight matrix. We search the best $\alpha$ that minimizes the loss function (20 candidates in toal) over the interval of [0, 1]. \cref{fig:quant} shows an overview of the injection and search. Since $\mathbf{I}_{m\ell}$ is precomputed before the search of $\alpha$, we go through every target module iteratively to finalize $\alpha$ and save the corresponding $\mathbf{s}$. The actual quantization is then performed by scaling the weights with $\mathbf{s}$ and applying the standard quantization function.

\section{Experiment Setup}

\subsection{Datasets}
\label{sec:datasets}
To demonstrate the effectiveness of \model{}, we select four reasoning benchmarks: AIME-120 \cite{MAAInvitationalCompetitions} for testing mathematical reasoning on 120 complex questions from 2022 to 2025, FOLIO \cite{han-etal-2024-folio} for testing logical reasoning, temporal sequences of BIG-Bench Hard \cite{suzgun2022challengingbigbenchtaskschainofthought} for testing temporal reasoning, and GPQA-Diamond \cite{rein2024gpqa} for testing scientific reasoning across multiple STEM domains. They cover a wide range of reasoning tasks with various difficulty levels. We measure accuracy of LRMs on each benchmark and report the average score (``Avg'' in tables). We use a combination of AIME 2021 \cite{aime_1983_2024} and 100 randomly sampled instances from the FOLIO training set as validation data.

Regarding calibration, we stick to the default calibration set of each baseline. We use the same calibration set for \model{} as in AWQ, and the calibration set of \model{} and AWQ is smaller than other baselines. We also collect verified reasoning chains \cite{guha2025openthoughtsdatarecipesreasoning} as the reasoning calibration data. This reasoning set is larger than AWQ's default but has the same size as other baselines. Since we find that quantization using activation does not benefit from the larger reasoning calibration (\cref{tab:motivation,tab:ablation}), we pursue the default calibration of each baseline for fair analysis. More details related to the smaller calibration set of \model{} and AWQ are in \cref{appendix:smaller_calibration}.

\subsection{Models and Pseudo-Fine-Tuning}
To show the wide applicability of \model{} over different model families and finetuning pipelines, we run \model{} on SFT, DPO, and RL fine-tuned LRMs, and also on a pre-fine-tuned LRM via pseudo-fine-tuning. For SFT models, we select a few R1 distillation LRMs \cite{Guo_2025}: \texttt{R1-Distill-Llama-70B}, \texttt{R1-Distill-Qwen-32B}, and \texttt{R1-0528-Qwen3-8B}. For the RL model, we select \texttt{Olmo-3-7B-Think} \cite{olmo2025olmo3}. It is fine-tuned via reinforcement learning from verifiable rewards (RLVR). We also select \texttt{Olmo-3-7B-Think-DPO} as the choice of DPO to diversity our selection. Since recent LRMs are typically fine-tuned through multiple stages, we leverage the weight updates from \texttt{Olmo-3-7B-Think-DPO} to the RL version to quantize the RL fine-tuned model and the updates from \texttt{Olmo-3-7B-Think-SFT} to the DPO version to quantize the DPO model.

When we need to quantize an LRM but its pre-fine-tuned checkpoint is not available (or it is not fine-tuned), we perform pseudo-fine-tuning on the target model to collect weight updates. In our experiment, we select \texttt{Qwen3-1.7B} \cite{yang2025qwen3technicalreport} as an example. It is worth noting that \texttt{Qwen3-1.7B} has its pre-fine-tuned version available, but its performance on hard reasoning tasks (\emph{e.g.}, AIME-120) is low. In this case, we assume that some fine-tuning is still needed, and we additionally train the model to collect fine-tuning signals for quantizing it.

\subsection{Baselines}
We compare only to PTQ, rather than other types such as QAT, since \model{} is a weight-only PTQ method with high efficiency (\cref{sec:speed}). We select all the PTQ methods discussed in \cref{sec:related_work} as baselines: GPTQ, GPTAQ, AWQ, and ANY3. They form a comprehensive list of state-of-the-art weight-only quantization methods. We do not report ANY3 on \texttt{R1-Llama-70B}, as its 3-bit quantized model is unable to generate coherent text. For DPO and RL fine-tuned LRMs, we report the performance of GPTQ and AWQ due to their overall competitive scores on SFT models.

\begin{table*}[t!]
\centering
\caption{Performance of 3-bit quantization on SFT, DPO, and RL fine-tuned LRMs. \model{} delivers a consistent improvement across various LRMs. We use the default calibration set for each PTQ method. \model{} reaches the best performance with one of the smallest calibration datasets. The highest scores of each model are \textbf{bold} excluding 16-bit.}
\label{tab:main_results}
\resizebox{\textwidth}{!}{%
\begin{tabular}{@{}cclccccc@{}}
\toprule
\multicolumn{3}{c}{Model Variants} & \multicolumn{5}{c}{Accuracy}  \\
\cmidrule(lr){1-3}\cmidrule(lr){4-8}
Model & Type & Precision & AIME-120  & FOLIO & Temporal & GPQA-Diamond  & Avg\\ \midrule
\texttt{R1-Llama-70B} & SFT & 16-bit & 56.7 & 79.8 & 100.0 & 60.6 & 74.28 \\
\texttt{R1-Llama-70B} & SFT & 3-bit GPTQ & 37.5 & 77.3 & 99.2 & \textbf{60.6} & 68.65 \\
\texttt{R1-Llama-70B} & SFT & 3-bit GPTAQ & 38.3 & 76.8 & 99.2 & 50.0 & 66.08 \\
\texttt{R1-Llama-70B} & SFT & 3-bit AWQ & 44.2 & 74.9 & \textbf{100.0} & 57.6 & 69.18\\
\texttt{R1-Llama-70B} & SFT & 3-bit \model{} & \textbf{49.2} & \textbf{78.3} & 99.6 & 58.1 & \textbf{71.30} \\
\midrule
\texttt{R1-Qwen-32B} & SFT & 16-bit & 56.7 & 83.7 & 99.2 & 61.1 & 75.18\\
\texttt{R1-Qwen-32B} & SFT & 3-bit GPTQ & 41.7 & 77.3 & \textbf{99.6} & 54.0 & 68.15\\
\texttt{R1-Qwen-32B} & SFT & 3-bit GPTAQ & 35.0	& 72.9	& \textbf{99.6}	& 60.1	& 66.90\\
\texttt{R1-Qwen-32B} & SFT & 3-bit AWQ & 50.0 & 77.3 & \textbf{99.6} & 55.1 & 70.50\\
\texttt{R1-Qwen-32B} & SFT & 3-bit ANY3 & 44.2 & 80.3 & \textbf{99.6} & 58.1 & 70.55\\
\texttt{R1-Qwen-32B} & SFT & 3-bit \model{} & \textbf{53.3} & \textbf{80.8} & \textbf{99.6} & \textbf{60.6} & \textbf{73.58}\\
\midrule
\texttt{R1-Qwen3-8B} & SFT & 16-bit & 59.2	& 79.8 & 99.6 & 63.6 & 75.55\\
\texttt{R1-Qwen3-8B} & SFT & 3-bit GPTQ & 8.3 & 77.3 & 82.8 & 35.9 & 51.08\\
\texttt{R1-Qwen3-8B} & SFT & 3-bit GPTAQ & 6.7 & \textbf{78.8} & 84.4 & 30.3 &	50.05\\
\texttt{R1-Qwen3-8B} & SFT & 3-bit AWQ & 34.2 & 77.8 & 98.4 &	47.5	& 64.48\\
\texttt{R1-Qwen3-8B} & SFT & 3-bit ANY3 & 15.0	& 69.0 &	54.4 &	30.3 & 42.18\\
\texttt{R1-Qwen3-8B} & SFT & 3-bit \model{} &  \textbf{40.0} & 76.8 & \textbf{99.2} & \textbf{48.5} & \textbf{66.13}\\
\midrule
\texttt{Olmo-3-7B-Think-DPO} & DPO & 16-bit &  57.5 & 70.4 & 98.8 & 49.0 & 68.93\\
\texttt{Olmo-3-7B-Think-DPO} & DPO & 3-bit GPTQ &  16.7 & 63.5 & 95.6 & \textbf{39.4} & 53.80\\
\texttt{Olmo-3-7B-Think-DPO} & DPO & 3-bit AWQ &  32.5 & \textbf{69.5} & 94.8 & 35.9 & 58.18 \\
\texttt{Olmo-3-7B-Think-DPO} & DPO & 3-bit \model{} &  \textbf{35.8} & 65.0 & \textbf{96.4} & 36.4 & \textbf{58.40} \\
\midrule
\texttt{Olmo-3-7B-Think} & RLVR & 16-bit &  54.2 & 75.4 & 99.6 & 45.5 & 68.68\\
\texttt{Olmo-3-7B-Think} & RLVR & 3-bit GPTQ &  21.7 & 69.5 & 93.6 & 41.9 & 56.68\\
\texttt{Olmo-3-7B-Think} & RLVR & 3-bit AWQ &  28.3 & \textbf{76.4} & 84.0 & 35.9 & 56.15\\
\texttt{Olmo-3-7B-Think} & RLVR & 3-bit \model{} &  \textbf{38.3} & 74.4 & \textbf{96.8} & \textbf{43.4} & \textbf{63.23}\\
\bottomrule
\end{tabular}%
}
\end{table*}

\subsection{Implementation Details}
\label{sec:implementation_details}
As we run \model{} on many LRMs (various model families with different sizes), model adaptation is needed. Since our pipeline uses only 16-bit arithmetic, large models can produce large $\mathbf{I}_{m\ell}^c$ (\emph{e.g.}, large matrices often yield more zero updates), leading to overflow; we mitigate this by slicing $\Delta \mathbf{W}$ of MLP modules into pieces (\emph{e.g.}, 2 or 5 pieces) when counting zero updates (used for \texttt{R1-Llama-70B} and \texttt{R1-Qwen-32B}). We then multiply the average zero-update count by the quadratic mapping scores in \cref{eq:final_I} to obtain the final channel importance. Another adaptation is to multiply our computed channel importance by the absolute value of activation, which helps on \texttt{R1-Llama-70B} but does not generalize, and we choose all adaptations based on their performance on our validation set.


We mainly focus on analyzing 3-bit quantization (W3A16) performance and also show 4-bit performance in \cref{sec:4_bit}, since 4-bit PTQ baselines have almost loseless scores \cite{zhang2025reasoningmeetscompressionunderstanding} and our priority is on 3-bit. When reporting accuracy scores of all methods, we use vLLM to perform inference. In \cref{sec:speed}, we first perform 4-bit quantization and then use the AWQ kernel to measure inference latency, since \model{} is compatible with AWQ kernel and the kernel supports only 4-bit LRMs.

\section{Results and Analysis}
\label{sec:results_analysis}
Our results aim to answer the following research questions:
\begin{itemize}[noitemsep,topsep=0pt,leftmargin=0.4cm]
    \item \textbf{RQ 1}: How does \model{} compare against other state-of-the-art PTQ baselines (\cref{sec:overall_results,sec:pseudo})?
    \item \textbf{RQ 2}: Is our design of \model{} optimal (\cref{sec:ablation})?
    \item \textbf{RQ 3}: Does \model{} support LRMs when their pre-fine-tuned checkpoints are not available (\cref{sec:pseudo})?
    \item \textbf{RQ 4}: Is \model{} an efficient method (\cref{sec:speed})?
    \item \textbf{RQ 5}: Is \model{} still competitive on 4-bit quantization (\cref{sec:4_bit})?
\end{itemize}

\begin{table*}[t!]
\caption{Ablation study. Performance of quantizing \texttt{R1-Distill-Qwen-32B}, \texttt{R1-Distill-Llama-8B}, and \texttt{R1-Qwen3-8B} to 3-bit while keeping a fraction of channels in 16-bit based on various protection signals (the top 5\% of weights by channel importance are protected in the 32B model, while the top 30\% are protected in the 8B LRMs).}
\label{tab:ablation}
\resizebox{\textwidth}{!}{%
\begin{tabular}{@{}clccccc@{}}
\toprule
\multicolumn{2}{c}{Model Variants} & \multicolumn{5}{c}{Accuracy}  \\
\cmidrule(lr){1-2}\cmidrule(lr){3-7}
Model & Protection Signal & AIME-120  & FOLIO & Temporal & GPQA-Diamond  & Avg\\ \midrule
\texttt{R1-Qwen-32B} & SFT both ends (no zero handling) &  34.2 & 71.4 & 98.4 & 47.5 & 62.88\\
\texttt{R1-Qwen-32B} & SFT both ends w/ 0s maximized &  30.8 & 76.4 & 99.2 & 50.5 & 64.23\\
\texttt{R1-Qwen-32B} & SFT both ends w/ 0s minimized &  32.5 & 74.9 & 98.8 & 48.5 & 63.68\\
\texttt{R1-Qwen-32B} & SFT both ends  &  \textbf{49.2} & \textbf{77.8} & \textbf{99.6} & \textbf{54.5} & \textbf{70.28}\\
\midrule
\texttt{R1-Llama-8B} & Activation on AWQ calibration & 5.0 & 52.7 & 28.0 & 19.7 & 26.35 \\
\texttt{R1-Llama-8B} & Activation on reasoning calibration & 5.0 & 49.3 & 27.6 & 14.1 & 24.00\\
\texttt{R1-Llama-8B} & SFT mid & 4.2 & 54.2 & 28.8 & 17.2 & 26.10\\
\texttt{R1-Llama-8B} & SFT both ends (no zero handling) & \textbf{20.0} & \textbf{63.5} & 56.8 & \textbf{36.4} & \textbf{44.18}\\
\texttt{R1-Llama-8B} & SFT both ends & 16.7 & 59.1 & \textbf{64.4} & 33.3 & 43.38 \\
\midrule
\texttt{R1-Qwen3-8B} & Activation on AWQ calibration & 0.0 & 39.4 & 24.8 & 11.1 & 18.83 \\
\texttt{R1-Qwen3-8B} & SFT both ends & \textbf{45.0} & \textbf{72.9} & \textbf{99.6} & \textbf{45.5} & \textbf{65.75}\\
\bottomrule
\end{tabular}%
}
\end{table*}

\subsection{Overall Results}
\label{sec:overall_results}
Our overall results of 3-bit quantization are in \cref{tab:main_results}. As specified in \cref{sec:datasets,appendix:smaller_calibration}, \model{} yields state-of-the-art performance on reasoning tasks with one of the smallest calibration datasets. As the primary focus of this paper is to push the boundary of ultra-low-bit quantization, we discuss the 4-bit quantization performance in \cref{sec:4_bit}.

\textbf{Improvement on SFT.} \model{} delivers consistent improvements over GPTQ, GPTAQ, AWQ, and ANY3 across all three SFT models, increasing the average score by up to 3.03\%. \model{} improves the average 3-bit score over the strongest PTQ baseline by 2.12\% on \texttt{R1-Llama-70B}. Even on \texttt{R1-Qwen3-8B}, which has far fewer parameters (and thus less room for improvement), \model{} improves the average score by 1.65\% over the best baseline. Since AIME-120 is the hardest task selected (largest performance drop after quantization), it is exciting to see that \model{} always leads on this benchmark. The strong performance of \model{} demonstrates the advantage of using SFT signals for quantization.

\textbf{Improvement on DPO and RL.} \model{} also delivers consistent improvements over the most competitive baselines on DPO and RL fine-tuned LRMs, increasing the average score by up to 6.55\%. This demonstrates that \model{} works on various types of fine-tuning. The fundamental reason is that the traces left by different fine-tuning strategies overlap significantly. The gain from \model{} is smallest on DPO, which we attribute to relatively weak fine-tuning signals. Similar to SFT models, \model{} leads on AIME-120.

\begin{table*}[t!]
\centering
\caption{Performance of 4-bit quantization on \texttt{R1-Llama-70B}. We use the default calibration set for each PTQ method. \model{} reaches the best performance with one of the smallest calibration datasets.}
\label{tab:4_bit_results}
\resizebox{\textwidth}{!}{%
\begin{tabular}{@{}cclccccc@{}}
\toprule
\multicolumn{3}{c}{Model Variants} & \multicolumn{5}{c}{Accuracy}  \\
\cmidrule(lr){1-3}\cmidrule(lr){4-8}
Model & Type & Precision & AIME-120  & FOLIO & Temporal & GPQA-Diamond  & Avg\\ \midrule
\texttt{R1-Llama-70B} & SFT & 16-bit & 56.7 & 79.8 & 100.0 & 60.6 & 74.28 \\
\texttt{R1-Llama-70B} & SFT & 4-bit GPTQ & 55.8 & 79.3	& 98.8	& 64.6 & 74.63 \\
\texttt{R1-Llama-70B} & SFT & 4-bit AWQ & 60.8 & 77.8 & 99.2 & 64.6 & 75.60 \\
\texttt{R1-Llama-70B} & SFT & 4-bit \model{} & 58.3 & 77.3 & 99.6 & 66.2 & 75.35 \\
\bottomrule
\end{tabular}%
}
\end{table*}

\subsection{Ablation Study}
\label{sec:ablation}
To precisely decode the effect of each component in \model{}, we conduct an ablation study using the same mixed-precision quantization setting as in \cref{sec:key_idea}. Testing different protection signals on more LRMs, we report the performance in \cref{tab:ablation}.

Scores on \texttt{R1-Qwen-32B} highlight the importance of explicitly handling zero updates. When protecting both ends without handling zero updates separately, we refer to \cref{eq:both_ends}, which yields the lowest average score. Then, instead of counting zero updates, we choose to directly assign them to $\mathbf{y}_{\text{max}}$ or $\mathbf{y}_{\text{min}}$ (default 10 or 1). Both choices provide modest gains, especially when zero updates are maximized. Our best design counts zero updates and excludes them when fitting the quadratic functions, resulting in a 5.95\% improvement in average score.

On the two 8B LRMs, “SFT both ends” yields much better scores than activation-based signals. On \texttt{R1-Llama-8B}, protecting intermediate magnitudes (“SFT mid”) is not helpful, which echoes our discussion in \cref{sec:key_idea} and reinforces our motivation for protecting both ends. Notably, on this model, protecting both ends without separately handling zero updates achieves the best average score, slightly outperforming counting zero updates. This contrast highlights the diversity of our LRMs: counting zero updates is the best design on the 32B model, and since it is only marginally worse on \texttt{R1-Llama-8B}, we adopt counting zero updates as our final design. This ablation study demonstrates the benefits of protecting weight updates on both ends and using the zero-update count to amplify channel importance.

\begin{figure}[ht]
  \begin{center}
    \centerline{\includegraphics[width=\columnwidth]{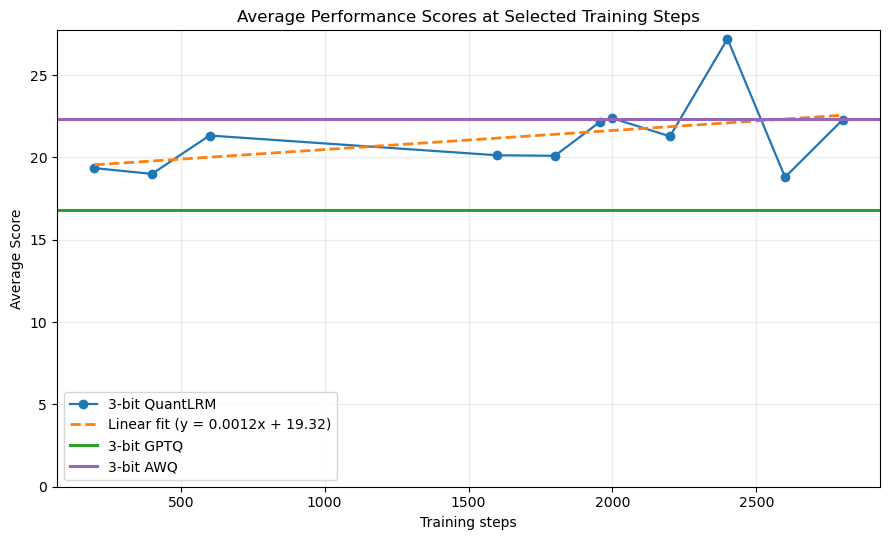}}
    \caption{Performance of 3-bit \model{} at different training steps. As training progresses, \model’s performance improves.}
    \label{fig:pseudo}
  \end{center}
\end{figure}

\subsection{Pseudo-Fine-Tuning}
\label{sec:pseudo}
A practical limitation of using fine-tuning traces is that a pre-fine-tuned checkpoint may be unavailable. Luckily, an exciting aspect of \model{} is that researchers can gather fine-tuning signals themselves when needed, which greatly enhances the applicability of our method. Because the ultimate goal is to quantize LRMs, fine-tuning can be stopped at any point once quantization performance is satisfactory. To show the results of quantizing a model via pseudo-fine-tuning, we select \texttt{Qwen3-1.7B} and perform SFT on the \texttt{OpenR1-Math-94k} dataset from LlamaFactory \cite{zheng2024llamafactory}. In \cref{fig:pseudo}, we gather weight updates at various training steps and measure the corresponding performance.

We see that \model{}’s performance gradually improves as training progresses, and this upward trend is also captured by our least-squares linear regression. The accumulation of weight updates as we have more training steps provides stronger and stronger fine-tuning signals. Since AWQ and GPTQ are applied directly to \texttt{Qwen3-1.7B} without training, their scores are shown as horizontal lines. Always surpassing GPTQ across all training steps, \model{} begins to match or surpass AWQ after 1956 training steps. We find that the fine-tuned \texttt{Qwen3-1.7B} has not yet reached convergence, as its 16-bit performance continues to improve after 2800 steps. Since it is not meaningful to quantize a model that has not yet converged, we only report the quantized \texttt{Qwen3-1.7B} via weight updates collected during pseudo-fine-tuning. Since \model{} benefits from stronger fine-tuning signals for successful quantization, we recommend running pseudo-fine-tuning for a few thousand training steps (or a few epochs).

\subsection{Cost of Time and Inference Speed}
\label{sec:speed}

\begin{table}[t!]
\centering
\caption{Cost of time and inference speed of \model{} and AWQ. We report the preparation time prior to quantization, the time used to search for scaling factors, and the inference latency of the 4-bit quantized model. \model{} adds little overhead for preparation and search, while achieving the same latency as AWQ.}
\resizebox{0.95\columnwidth}{!}{%
\begin{tabular}{lccc}
\toprule
Method & Prep Time & Search Time & Tokens/s \\ \midrule
AWQ & - & 6min28s &  30.24 \\ 
\model{} & 2min27s & 6min35s & 30.33 \\ 
\bottomrule
\end{tabular}%
}
\label{tab:speed}
\end{table}

As discussed in \cref{sec:our_method}, our method requires computing weight updates, mapping them via quadratic functions, and counting zero updates as preparation prior to quantization. During quantization, the core is to search for optimal scaling factors, so we record the search time of \model{} and AWQ. We also measure inference latency by reporting the median latency over seven prompts, as specified in \cref{appendix:prompts}. As mentioned in \cref{sec:implementation_details}, this experiment is conducted on 4-bit quantized \texttt{Olmo-3-7B-Think}.

In \cref{tab:speed}, we observe that \model{} incurs a small one-time preparation overhead of only 2 minutes and 27 seconds, which is not present in AWQ because AWQ relies purely on activation statistics. Importantly, this extra cost is \emph{offline} (prior to quantization) and does not affect inference. During quantization, both methods take roughly the same amount of time, with \model{} spending a few extra seconds loading the precomputed fine-tuning signals. Their inference latency is the same, since we use the same AWQ kernel for inference. Besides the cost of time, \model{} consumes similar amount of GPU memory as AWQ as well.

In conclusion, paying a small amount of extra cost mainly during preparation, \model{} is a reasonably efficient method that delivers significantly better reasoning accuracy after quantization.

\subsection{Performance of W4A16 Quantization}
\label{sec:4_bit}

We report performance scores of W4A16 in \cref{tab:4_bit_results}. The average scores of all three methods surpass the 16-bit model, which echoes recent analyses \cite{zhang2025reasoningmeetscompressionunderstanding,liu2025quantization}. Even on AIME-120 and GPQA-Diamond (challenging benchmarks), the 16-bit \texttt{R1-Llama-70B} ranks third or worse. The is the reason why we primarily focus on the performance of W3A16, as 4-bit LRMs perform similarly with little room of improvement. As discussed in \cref{sec:overall_results}, \model{} pushes the boundary of 3-bit quantization and delivers state-of-the-art performance.

From a practical perspective, these findings indicate that \model{} serves as a “safe default” across bit-widths: it provides substantial improvements in W3A16 while remaining competitive and stable in W4A16, where current PTQ methods already saturate in performance.

\section{Conclusion}
In this paper, we propose \model{}, which stands for quantization of LRMs via fine-tuning signals. \model{} gathers weight updates during SFT, DPO, and RL fine-tuning and assigns high importance scores to protect the smallest and largest updates. We empirically demonstrate that \model{} delivers a consistent improvement for LRMs quantization. Compatible with vLLM and AWQ kernel, \model{} offers great inference flexibility with comparable speedup to state-of-the-art quantization methods. More broadly, our findings suggest that fine-tuning traces provide a strong and lightweight importance signal for PTQ, offering a practical direction to future research.

\clearpage







\section*{Impact Statement}

This paper presents work whose goal is to advance the field of Machine
Learning. There are many potential societal consequences of our work, none
which we feel must be specifically highlighted here.

\nocite{langley00}

\bibliography{example_paper}
\bibliographystyle{icml2026}

\newpage
\appendix
\onecolumn
\section{A Smaller Calibration Set of \model{} and AWQ}
\label{appendix:smaller_calibration}
We use the default calibration size for all quantization methods. For GPTQ, GPTAQ, and ANY3, we use 128 samples of length 2048 tokens from the C4 dataset \cite{raffel2023exploringlimitstransferlearning}. For AWQ and \model, we use 128 samples of length 512 instead. Note that \model{} still outperforms other baselines despite using one of the smallest calibration sets, demonstrating the efficacy of our method.

\section{Prompts Used for Speed Analysis}
\label{appendix:prompts}
We use 7 short prompts to measure the inference speed of \model{} and AWQ:
\begin{itemize}[noitemsep,topsep=0pt,leftmargin=0.4cm]
    \item ``\textit{Deeply analyze; then decide.}''
    \item ``\textit{Enumerate approaches; pick best.}''
    \item ``\textit{Solve; stress-test; revise.}''
    \item ``\textit{Assume nothing; derive everything.}''
    \item ``\textit{Answer; justify; quantify uncertainty.}''
    \item ``\textit{Decompose; solve parts; recombine.}''
    \item ``\textit{List risks; add safeguards; finalize.}''
\end{itemize}

The goal is to generate 200 reasoning tokens for each prompt and compute the median latency.

\end{document}